\documentclass[conference]{IEEEtran}
\IEEEoverridecommandlockouts

\usepackage{cite}
\usepackage{amsmath,amssymb,amsfonts}
\usepackage{algorithmic}
\usepackage{graphicx}
\usepackage{textcomp}
\usepackage{xcolor}
\def\BibTeX{{\rm B\kern-.05em{\sc i\kern-.025em b}\kern-.08em
    T\kern-.1667em\lower.7ex\hbox{E}\kern-.125emX}}
\begin{document}

\title{Neural Video Representation for Redundancy Reduction and Consistency Preservation
}

\author{\IEEEauthorblockN{Taiga Hayami}
\IEEEauthorblockA{\textit{Graduate School of FSE,} \\
\textit{Waseda University}\\
Tokyo, Japan \\
hayatai17@fuji.waseda.jp}
\and
\IEEEauthorblockN{Takahiro Shindo}
\IEEEauthorblockA{\textit{Graduate School of FSE,} \\
\textit{Waseda University}\\
Tokyo, Japan \\
taka\_s0265@ruri.waseda.jp}
\and
\IEEEauthorblockN{Shunsuke Akamatsu}
\IEEEauthorblockA{\textit{Graduate School of FSE,} \\
\textit{Waseda University}\\
Tokyo, Japan \\
s.akamatsu@akane.waseda.jp}
\and
\IEEEauthorblockN{Hiroshi Watanabe}
\IEEEauthorblockA{\textit{Graduate School of FSE,} \\
\textit{Waseda University}\\
Tokyo, Japan \\
hiroshi.watanabe@waseda.jp}
}

\maketitle

\begin{abstract}
Implicit neural representation (INR) embed various signals into neural networks.
They have gained attention in recent years because of their versatility in handling diverse signal types.
In the context of video, INR achieves video compression by embedding video signals directly into networks and compressing them.
Conventional methods either use an index that expresses the time of the frame or features extracted from individual frames as network inputs.
The latter method provides greater expressive capability as the input is specific to each video.
However, the features extracted from frames often contain redundancy, which contradicts the purpose of video compression.
Additionally, such redundancies make it challenging to accurately reconstruct high-frequency components in the frames.
To address these problems, we focus on separating the high-frequency and low-frequency components of the reconstructed frame.
We propose a video representation method that generates both the high-frequency and low-frequency components of the frame, using features extracted from the high-frequency components and temporal information, respectively.
Experimental results demonstrate that our method outperforms the existing HNeRV method, achieving superior results in 96 percent of the videos.

\end{abstract}


\section{Introduction}
The demand for high-quality video has surged, driven by the widespread availability of the internet and the proliferation of video streaming services. 
Video compression technology is essential for the efficient transmission and storage of such large amounts of video data.
While video compression research has a long history, the Moving Picture Experts Group (MPEG) has standardized many key compression techniques. 
Traditional video coding standards, such as HEVC/H.265 \cite{hevc} and VVC/H.266 \cite{vvc}, rely on manually crafted, rule-based algorithms. 
In recent years, with the development of deep learning, neural network-based video compression methods have been actively studied.
Among these, approaches that embed  video signals directly into neural networks have garnered significant attention. 

\begin{figure}[tb]
\centerline{\includegraphics[width=1.0\columnwidth]{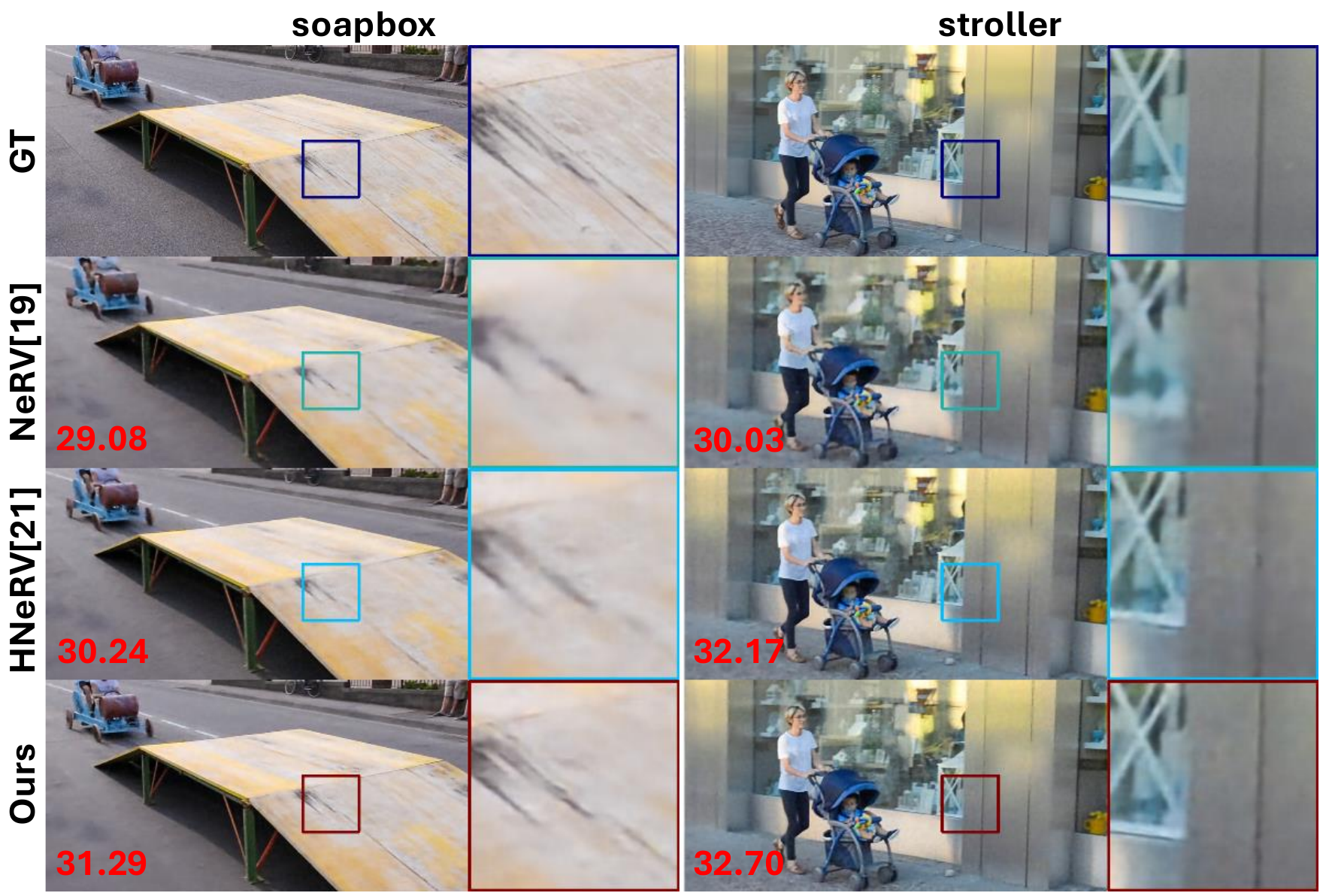}}
\caption{Visualization of reconstruction results in ``soapbpx'' and ``stroller'' videos. The red number is the PNSR for the frame.}
\label{result}
\end{figure}

\begin{figure*}[tb]
\centerline{\includegraphics[width=2.0\columnwidth]{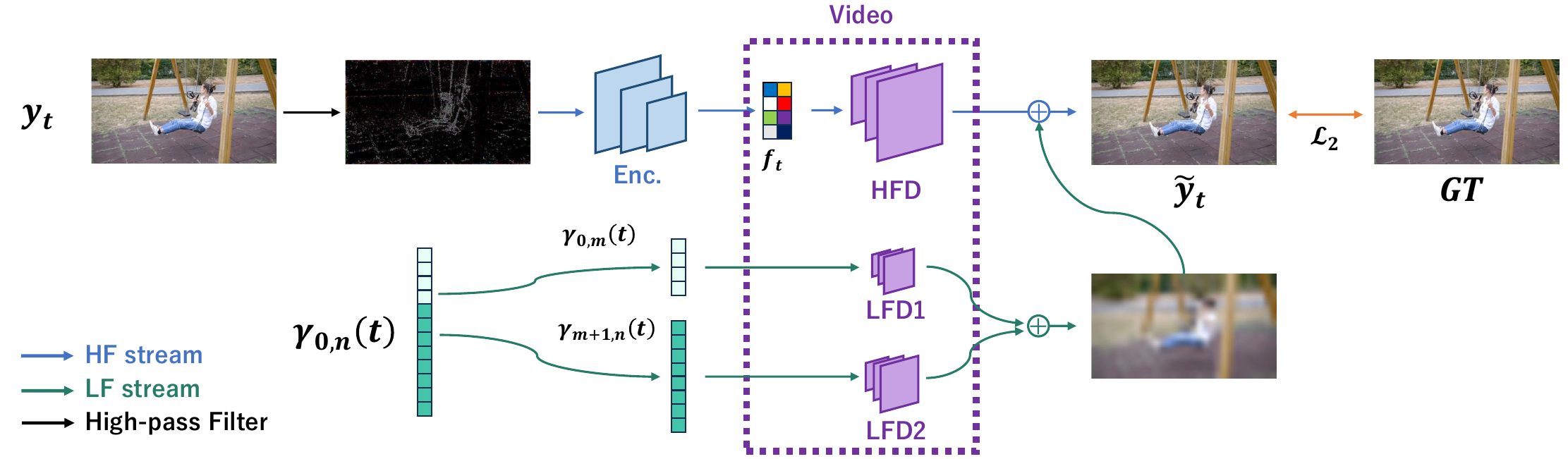}}
\caption{The architecture of proposed method. It consists of the HF-stream (blue arrow) and the LF-stream (green line). The part circled in purple is treated as video.}
\label{pipeline}
\end{figure*}

Implicit neural representation (INR) overfits various signals into neural networks.
This method allows for the compact representation of even complex signals.
For instance, INR-based methods such as Neural Radiance Fields (NeRF) \cite{nerf}, along with its variants \cite{mipnerf, enerf, regnerf, refsr-nerf}, have gained attention as novel approaches for 3D scene representation. 
This signal representation approach extends to images\cite{siren, coin, coinpp, coolchic, c3}, videos\cite{nerv, enerv, hnerv, dnerv, dsnerv}, and 3D shapes\cite{shape, shape2, shape3}.
Since the network that embeds the signal can be regarded as the signal itself, compressing that network equals to coding the signal.
Specifically, in the case of videos, INR enables video compression by embedding and subsequently compressing the neural network.

Compared to conventional video compression methods, which have complex pipelines, INRs reduce the computational cost of decoding by utilizing a simpler network structure. 
Neural Representations for Videos (NeRV) \cite{nerv} encode videos by using a network that takes frame time indices as inputs and outputs the corresponding frames. 
These indexes are independent of the video content.
Therefore, the quality of the reconstructed frames relies solely on the network's ability to represent the video.
In contrast, HNeRV\cite{hnerv} uses features extracted from each frame of the video as inputs to the network.
Using inputs that capture the unique characteristics of each video allows HNeRV to provide a generalized representation.
Consequently, the quality of the reconstructed frames depends not only on the network's representational capacity but also on the content of the input features. 
The total information required to represent a video is the sum of the network’s information content and the content of all input features.

However, the input features used by HNeRV introduce redundancy, as the feature values between adjacent frames are often nearly identical. 
In video compression, reducing redundant information is essential to improving efficiency and the quality of the reconstructed frames. 
To address this, the input features must be carefully designed. 
Due to the spectral bias inherent in deep learning \cite{sb, sb2, sb3, finer}, where networks tend to learn low-frequency components more easily than high-frequency ones, it is also difficult to reconstruct high-frequency details in frames with similar inputs.
In addition, since this method does not use temporal information as inputs, it is unclear whether the network effectively understands the relationship between frames in the video.

In this paper, we propose a method that separates the high-frequency and low-frequency components of each frame, leveraging both frame-specific and temporal information.
The high-frequency components of the video have little temporal consistency, and the features extracted from them show different characteristics from each other.
This approach effectively reduces input redundancy. 
Meanwhile, the temporally consistent low-frequency components are reconstructed using temporal information. 
By combining these two approaches, our method improves video representation and compression performance (as illustrated in Fig. \ref{result}).

\section{Related Works}

\subsection{Implicit Neural Representation for Videos}
In recent years, Implicit Neural Representation (INR) have been applied to a wide range of tasks. 
Most tasks, such as image and scene representation, employ coordinate-based INR \cite{siren, coin, coinpp, coolchic, c3, nerf, mipnerf, enerf, regnerf, refsr-nerf}. 
For instance, the INR-based method COIN\cite{coin} estimates pixel color based on its position $(x, y)$. 
In the context of video, this approach extends to use both pixel position and frame time $(x, y, t)$. 
However, as the number and size of frames increase, the number of input pairs grows substantially, leading to longer times required for training and inference.
 Due to these limitations, frame-based approaches like NeRV \cite{nerv} have gained more attention compared to coordinate-based methods.

Frame-based methods map frames from specific inputs.
NeRV introduced an index-based structure where the input is a frame's time index, and the output is the corresponding frame.
Since this input type is independent of frame size, the training time is relatively short compared to coordinate-based approaches. 
ENeRV \cite{enerv} enhances performance by incorporating spatial coordinates in addition to frame indices. 
DS-NeRV \cite{dsnerv} further enhances reconstruction quality by separating the dynamic and static components of the video. 
In contrast, HNeRV \cite{hnerv} employs a hybrid-based structure that uses the features extracted from each frame as inputs to the network. 
This method captures video-specific patterns more effectively than index-based approaches, which are agnostic to individual video content. 
DNeRV \cite{dnerv} extends the hybrid-based approach by integrating difference images from neighboring frames, resulting in a more robust video representation.

\begin{figure}[tb]
\centerline{\includegraphics[width=0.8\columnwidth]{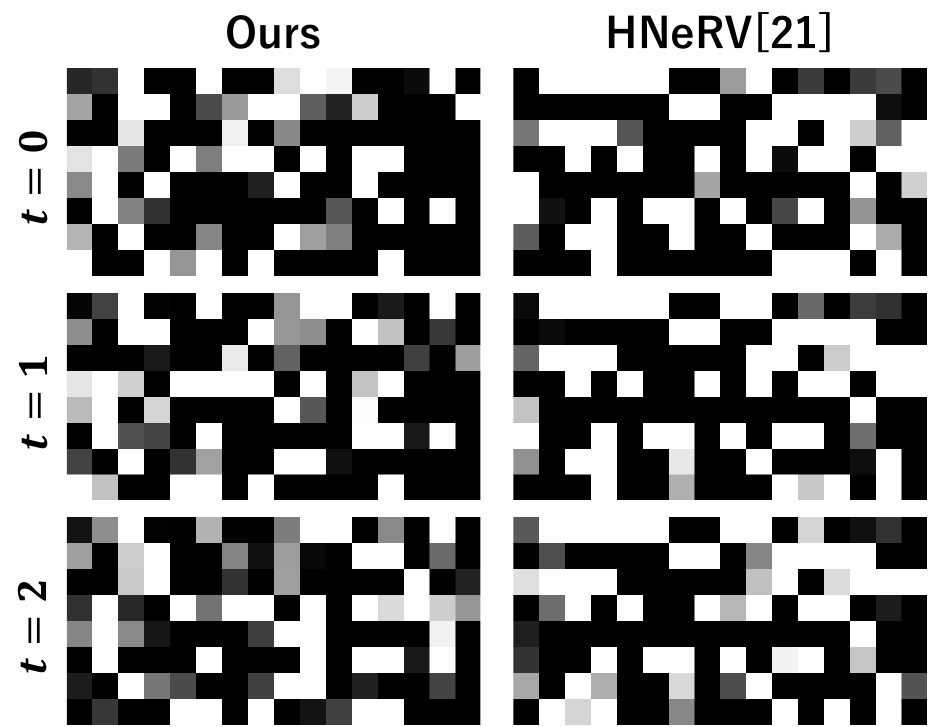}}
\caption{Visualization results of features extracted by the encoder.}
\label{feature}
\end{figure}

\begin{figure*}[tb]
\centerline{\includegraphics[width=2.0\columnwidth]{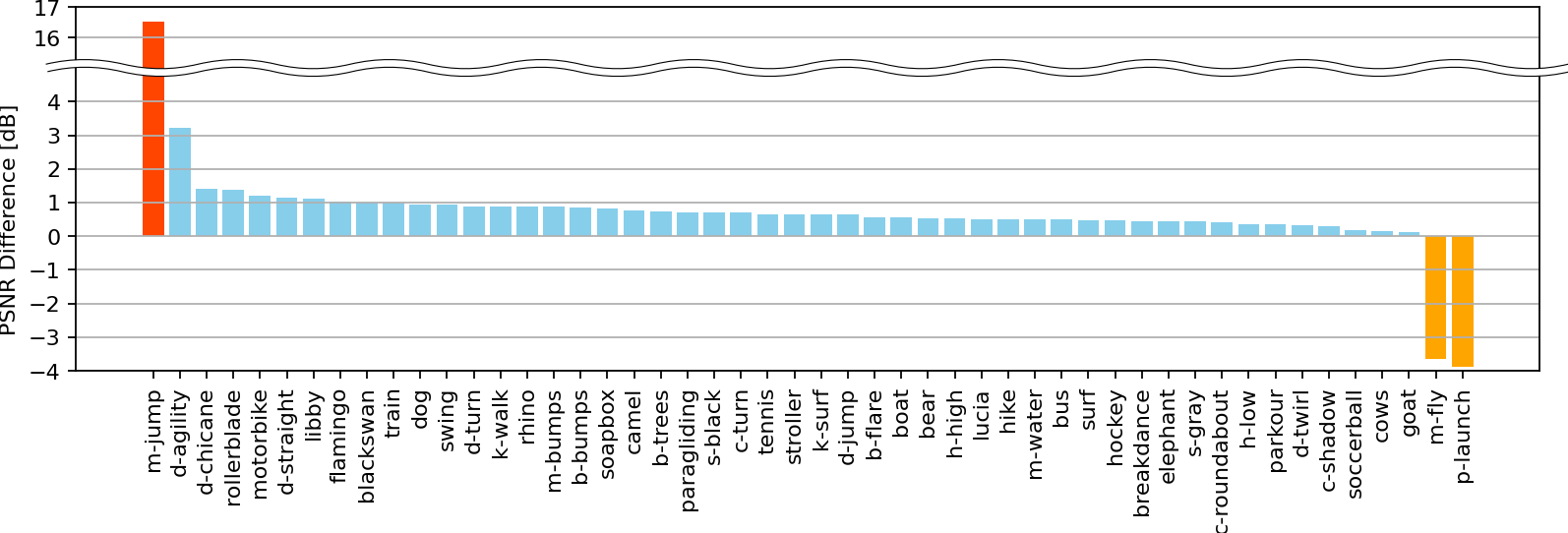}}
\caption{Comparison of PSNR between the proposed method and HNeRV for reconstructed videos. 
The horizontal axis represents the video sequences from the DAVIS dataset, and the vertical axis represents the PSNR difference between the proposed method and HNeRV. 
Positive values on the vertical axis indicate that the proposed method outperforms HNeRV. 
Video sequences where HNeRV did not perform well are highlighted in red, while those using the proposed method are highlighted in orange. 
The proposed method ensures a minimum level of quality even when training does not fully converge.
}
\label{vs}
\end{figure*}

\subsection{Video Compression}
Conventional video coding standards, such as HEVC/H.265 \cite{hevc} and VVC/H.266 \cite{vvc}, operate based on predefined rules and algorithms to achieve efficient and fast compression. 
In recent years, deep learning-based video compression methods have been actively studied \cite{dvc, dcvc}.  
While these learning-based methods offer improved compression efficiency over traditional rule-based methods, they come with higher training costs and slower decoding speeds due to their more complex structures.

INR-based methods primarily focus on efficiently embedding video data into a neural network. 
These approaches achieve video compression by applying techniques such as branch pruning, quantization, and entropy coding to the network weights where the video is embedded. 
The simplicity of this structure allows for fast decoding and optimized compression tailored to each specific video. 
However, this method lack versatility, as embedding and compression must be trained separately for each video. 
NVRC \cite{nvrc} focused on improving compression methods in INR-based video compression. 
They proposed a framework that integrates entropy coding and quantization models, achieving performance that surpasses VVC.

\section{Proposed Method}
\subsection{Overview}
In video compression, the efficient use of a limited amount of information to represent the video is crucial.
Similarly, efficient video embedding is critical for INR for videos. 
However, the spectral bias problem \cite{sb, sb2, sb3, finer} makes it difficult for networks to accurately reconstruct the high-frequency components of frames. 
To address this, we propose a video representation method that leverages both the high-frequency components of frames and temporal information, as shown in Figure \ref{pipeline}. 
The proposed method is divided into two streams: a high-frequency component stream (HF-stream) and a low-frequency component stream (LF-stream), each responsible for reconstructing their respective components of the frame.
The two streams are linked in a manner that serves the same role as residual connections \cite{resnet, rchnerv}.

\subsection{HF-stream}
The network responsible for reconstructing the high-frequency component (HFD) uses a hybrid-based structure and takes frame-specific information as input. 
The feature size in this structure is very small (e.g., $16 \times 2 \times 4$), making efficient feature extraction essential. 
In HNeRV, features are extracted from the entire frame, which introduces redundancy, as adjacent frames tend to be similar and therefore produce similar features. 
As shown in Fig \ref{feature}, the $16 \times 2 \times 4$ features extracted from adjacent frames are reshaped to $8 \times 16$ and visualized, respectively. 
This reveals that HNeRV generates similar features for adjacent frames. 
Consequently, when a frame is reconstructed from these similar features, the high-frequency components, which differ between adjacent frames, are likely to be lost. 

To address this issue, we extract features specifically from the high-frequency components of the frames. 
High-frequency components exhibit low temporal consistency, leading to minimal similarity between adjacent features.
This helps reduce feature redundancy and emphasizes finer details of each frame. 
The high-frequency components are obtained via a high-pass filter, and their corresponding features are extracted by an encoder. 
Similar to HNeRV, the encoder responsible for feature extraction utilizes ConvNeXt blocks \cite{convnext}, while the HFD is composed of multiple HNeRV blocks \cite{hnerv}.

\begin{table}[tb]
\caption{Comparison of the evaluation result averages of the reconstructed video for each method}
\begin{center}
\begin{tabular}{c|ccc}
\hline
Method                 & PSNR $\uparrow$   & MS-SSIM $\uparrow$ & LPIPS $\downarrow$ \\ \hline
NeRV(All)\cite{nerv}   & 28.91             & 0.8865             & 0.3663             \\ 
HNeRV(All)\cite{hnerv} & 30.94             & 0.9160             & 0.2955             \\ 
Ours(All)              & \textbf{31.81}    & \textbf{0.9284}    & \textbf{0.2780}    \\ \hline
HNeRV(Suc)\cite{hnerv} & 31.19             & 0.9261             & 0.2873             \\ 
Ours(Suc)              & \textbf{31.91}    & \textbf{0.9361}    & \textbf{0.26737}    \\ \hline
\end{tabular}
\end{center}
\label{compare}
\end{table}

\begin{figure*}[tb]
\centerline{\includegraphics[width=2.0\columnwidth]{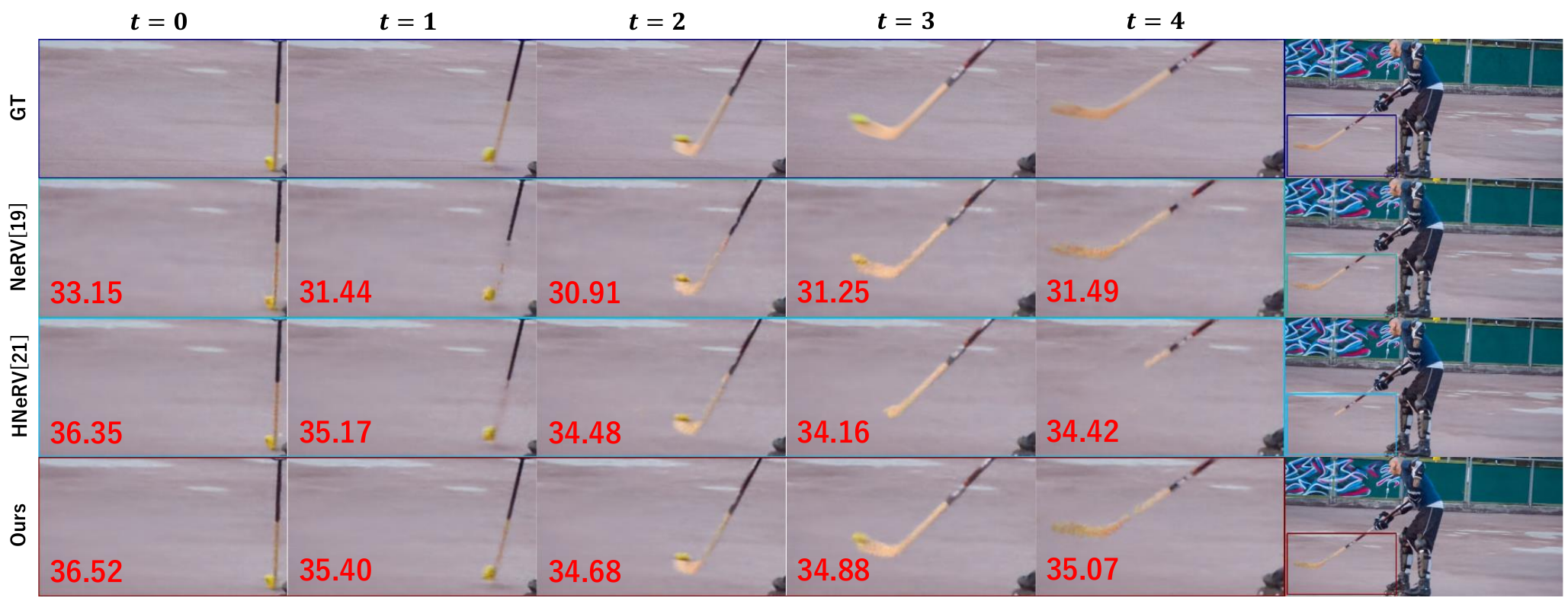}}
\caption{Visualization of consecutive frames from ``hockey'' video. The red numbers indicate the PSNR for the entire frame.}
\label{hockey}
\end{figure*}

\subsection{LF-stream}
The network responsible for reconstructing the low-frequency components employs an index-based structure, composed of a Multi-layer perceptron (MLP) and several NeRV blocks \cite{nerv}. 
To reduce the number of network parameters, the network is divided into two parts, LFD1 and LFD2. 
Low-frequency components are reconstructed using a frame-time index due to their high temporal relevance. 
This indexing approach is more efficient than using frame-specific information, as it does not increase the overall size of the video representation.

The index $t$ is normalized between (0, 1] and extended dimensionally using Position Encoding \cite{nerf, transformer, nerv}.
The Positional Encoding $\gamma$ is defined as follows,
\begin{equation}
\gamma_{0, n}(t) = (\sin(b^0\pi t), \cos(b^0\pi t), \cdots, \sin(b^n\pi t), \cos(b^n\pi t)),
\end{equation}
where $b$ and $n$ are hyper-parameters.
Each element of $\gamma_{0, n}(t)$ has a different sensitivity range depending on the value of $b^x$. 
Smaller values of $b^x$ are more sensitive to long-term changes in the video, while larger values are more responsive to short-term changes.
When $\gamma_{0, n}(t)$ with large $n$ is used as input to the MLP, the network has large parameters.
To mitigate this, $\gamma_{0, n}(t)$ is divided into two parts, $\gamma_{0, m}$ and $\gamma_{m+1, n}$, using the hyper-parameter $m$ ($0<m<n$), which serve as inputs to LFD1 and LFD2, respectively.
The number of MLP parameters is determined by its input and output dimensions, and dividing the network reduces the total parameter count. 
For instance, if $m$ is set to half of $n$, the number of MLP parameters is reduced by half. 
Since $\gamma$ plays different roles depending on the value of $b^x$, this division has minimal impact on overall performance.

\section{Experiment}

\begin{figure*}[tb]
\centerline{\includegraphics[width=2.0\columnwidth]{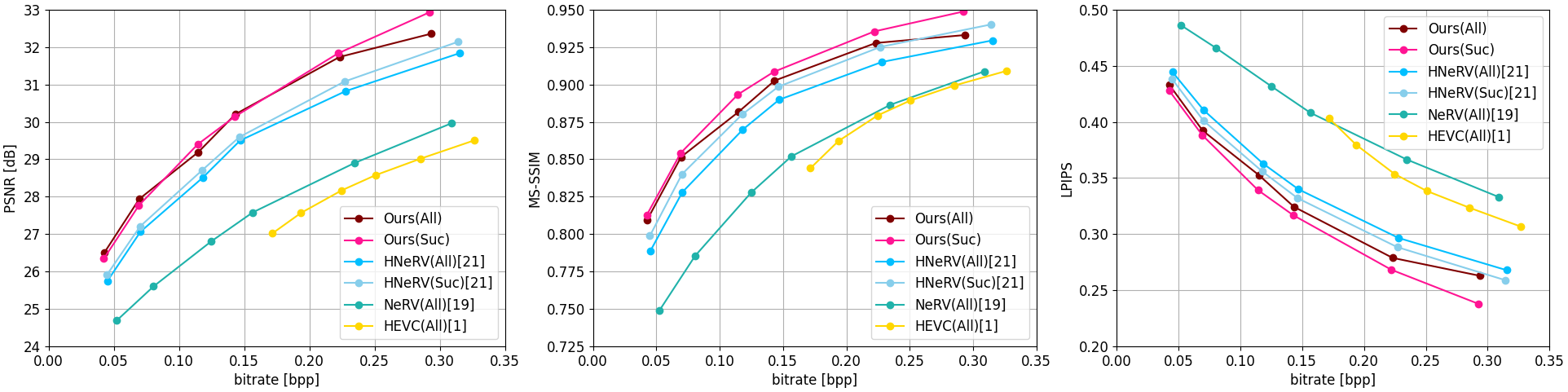}}
\caption{Video compression result on DAVIS dataset.}
\label{rd}
\end{figure*}

We use videos from the DAVIS \cite{davis} dataset, which includes $50$ videos with a resolution of $1080 \times 1920$ and frame counts ranging from $25$ to $104$. 
The frames are cropped to a size of $640 \times 1280$. 
To extract high-frequency components, we apply a high-pass filter that removes $80\%$ of the frequency bandwidth. 
The training process is conducted over $300$ epochs. 
We adjust the total number of features and decoder parameters to approximate $1.5$ million parameters as closely as possible, with the model sizes of HFD, LFD1, and LFD2 set in a ratio of $20:1:5$.
The hyper-parameters $b$, $m$, and $n$ in the Positional Encoding are set to $1.25$, $10$, and $30$, respectively.
The loss function is based on the L2 loss between the reconstructed and ground truth frames. 
For video compression, we apply quantization to the feature and decoder parameters (HFD, LFD1, LFD2) and use Huffman coding to compress the quantized weights, similar to HNeRV, with quantization factors set to $6$ and $8$, respectively.
We evaluate our method using Peak Signal-to-Noise Ratio (PSNR), Multi-Scale Structural Similarity (MS-SSIM), and Learned Perceptual Image Patch Similarity (LPIPS).

A comparison of PSNR between HNeRV and the proposed method for each video sequence is shown in Fig. \ref{vs}.
As discussed in PNeRV\cite{pnerv}, with the hybrid-base structure, HNeRV and the proposed method did not converge in learning for several videos.
In Fig. \ref{vs}, video sequences where HNeRV did not achieve adequate learning are highlighted in red, while those for the proposed method are shown in orange. 
For the "m-jump" video, HNeRV yielded a PSNR of 14.36 dB, indicating that learning did not converge. 
In contrast, the proposed method, which utilizes three decoders, efficiently learns LFD1 and LFD2 with a small model size. 
This strategy ensures a minimum quality even when learning in the HFD pathway does not converge. 
Except for these specific sequences, the proposed method outperforms HNeRV.
The evaluation results of the video reconstruction are shown in Table \ref{compare}.
This table shows the evaluation average of the reconstructed videos for each evaluation metric.
In addition, the hybrid-base structure of HNeRV and the proposed method show the evaluation averages only for the videos that were successfully learned (except for ``m-jump'', ``m-fly'', and ``p-launch'' video).

An example of consecutive frame visualizations from the``hockey'' video is shown in Fig. \ref{hockey}. 
The red numbers denote the PSNR for each entire frame. 
NeRV exhibits low overall quality, while HNeRV suffers from missing objects, such as sticks and balls, in several frames, failing to consistently reconstruct the video. 
In contrast, the proposed method consistently reconstructs detailed objects across the entire video, maintaining high-quality results. 
Additional visualization examples are provided in Fig. \ref{result}, further demonstrating the effectiveness of the proposed method in generating more intricate representations. 
For instance, fine details such as the board patterns in the ``soapbox'' video and the delicate lines in the ``stroller'' video are faithfully rendered. 
By extracting features from high-frequency components, as shown in Fig. \ref{feature}, the proposed method yields features with greater variation, contributing to its ability to capture fine details.

The video compression results on the DAVIS dataset are shown in Fig. \ref{rd}. 
Feature and model compression are applied across six model sizes ($0.3$M, $0.5$M, $0.8$M, $1.0$M, $1.5$M, $2.0$M). 
The proposed method demonstrates superior compression performance compared to HEVC and conventional INR-based methods.

\section{Conclusion}
In this paper, we proposed an INR-based video representation method that utilizes two streams to incorporate both frame and temporal information for reconstructing detailed frame representations. 
The former stream focuses on reconstructing the high-frequency components of the frame, while the latter stream handles the low-frequency components. 
The frame information specifically leverages high-frequency components to minimize redundancy.
Additionally, the network responsible for processing low-frequency components is divided to reduce the number of parameters. 
These approaches demonstrates strong performance in both video representation and compression. 
Future work will focus on optimizing hyper-parameter settings that are sensitive to individual video sequences.
Additionally, it is essential to investigate the underlying causes of learning collapse in certain video sequences.

\vspace{12pt}

\begin{thebibliography}{00}
\bibitem{hevc} High Efficiency Video Coding, Standard ISO/IEC 23008-2, ISO/IEC JTC 1, Apr.2013.
\bibitem{vvc} Versatile Video Coding, Standard ISO/IEC 23090-3, ISO/IEC JTC 1, Jul. 2020.
\bibitem{dvc} G. Lu, W. Ouyang, D. Xu, X. Zhang, C. Cai, and Z. Gao, ``DVC: An End-to-end Deep Video Compression Framework,'' IEEE/CVF Conference on Computer Vision and Pattern Recognition (CVPR), 11006-11015, Jun. 2019.
\bibitem{dcvc} J. Li, B. Li, and Y. Lu, ``Deep Contextual Video Compression,'' Advances in Neural Information Processing Systems  (NeurIPS), Dec. 2021.
\bibitem{nerf} B. Mildenhall, P. Srinivasan, M. Tancik, J. Barron, R. Ramamoorthi, and R. Ng, ``NeRF: Representing Scenes as Neural Radiance Fields for View Synthesis,'' European Conference on Computer Vision (ECCV), 405-421, Nov. 2020.
\bibitem{mipnerf} J. Barron, B. Mildenhall, M. Tancik, P. Hedman, R. Martin-Brualla, and P. Srinivasan, ``Mip-NeRF: A Multiscale Representation for Anti-Aliasing Neural Radiance Fields,'' IEEE/CVF International Conference on Computer Vision (ICCV), 5855-5864, Oct. 2021.
\bibitem{enerf} T. Hu, S. Liu, Y. Chen, T. Shen, and J. Jia, ``EfficientNeRF  Efficient Neural Radiance Fields,'' IEEE/CVF Conference on Computer Vision and Pattern Recognition (CVPR), 12902-12911, Jun. 2022.
\bibitem{regnerf} M. Niemeyer, J. Barron, B. Mildenhall, M. Sajjadi, A. Geiger, and N. Radwan, ``RegNeRF: Regularizing Neural Radiance Fields for View Synthesis From Sparse Inputs,'' IEEE/CVF Conference on Computer Vision and Pattern Recognition (CVPR), 5480-5490, Jun. 2022.
\bibitem{refsr-nerf} X. Huang, W. Li, J. Hu, H. Chen, and Y. Wang, ``RefSR-NeRF: Towards High Fidelity and Super Resolution View Synthesis,'' IEEE/CVF Conference on Computer Vision and Pattern Recognition (CVPR), 8244-8253, Jun. 2023.
\bibitem{siren} S. Vincent, M. Julien, B. Alexander, L. David, and W. Gordon, ``Implicit Neural Representations with Periodic Activation Functions,'' Advances in Neural Information Processing Systems (NeurIPS), 7462-7473, 2020.
\bibitem{coin} D. Emilien, G. Adam, A. Milad, T. Yee, and D. Arnaud, ``COIN: COmpression with Implicit Neural representations,'' arXiv preprint arXiv:2103.03123, 2021.
\bibitem{coinpp} D. Emilien, L. Hrushikesh, A. Milad, G. Adam, T. Yee, and D. Arnaud, ``COIN++: Neural Compression Across Modalities,'' arXiv preprint arXiv:2201.12904, 2022. 
\bibitem{coolchic} T. Ladune, P. Philippe, F. Henry, G. Clare, and T. Leguay, ``COOL-CHIC: Coordinate-based Low Complexity Hierarchical Image Codec,'' IEEE/CVF International Conference on Computer Vision (ICCV), 13515-13522, Oct. 2023.
\bibitem{c3} H. Kim, M. Bauer, L. Thesis, J. Schwarz, and E. Dupont, ``C3: High-Performance and Low-Complexity Neural Compression from a Single Image or Video,'' IEEE/CVF Conference on Computer Vision and Pattern Recognition (CVPR), 9347-9358, Jun. 2024.
\bibitem{shape} K. Genova, F. Cole, D. Vlasic, A. Sarna, W. Freeman, and T. Funkhouser, ``Learning Shape Templates With Structured Implicit Functions,'' IEEE/CVF International Conference on Computer Vision (ICCV), 7154-7164, Oct. 2019.
\bibitem{shape2} Z. Chen, H. Zhang, ``Learning Implicit Fields for Generative Shape Modeling,'' IEEE/CVF Conference on Computer Vision and Pattern Recognition (CVPR), 5939-5948, Jun. 2019.
\bibitem{shape3} L. Luca, C. Adriano, S. Riccardo, R. Pierluigi, S. Samuele, and S. Luigi, ``Deep Learning on Implicit Neural Representations of Shapes,'' arXiv preprint arXiv:2302.05438, 2023.
\bibitem{transformer} A. Vaswani, N. Shazeer, N. Parmar, J. Uszkoreit, L. Jones, A. Gomez, L. Kaiser, and I. Polosukhin, ``Attention is All you Need,'' Advances in Neural Information Processing Systems (NeurIPS), 5998-6008, Dec. 2017.
\bibitem{nerv} H. Chen, B. He, H. Wang, Y. Ren, S. Lim, and A. Shrivastava, ``NeRV: Neural Representations for Videos,'' Advances in Neural Information Processing Systems (NeurIPS), 21557-21568, Nov. 2021.
\bibitem{enerv} L. Zizhang, W. Mengmeng, P. Huaijin, X. Kechun, M. Jianbiao, and L. Yong, ``E-NeRV: Expedite Neural Video Representation with Disentangled Spatial-Temporal Context,'' European Conference on Computer Vision (ECCV), 267-284, Nov. 2022.
\bibitem{hnerv} H. Chen, M. Gwilliam, S. Lim, and A. Shrivastava, ``HNeRV: A Hybrid Neural Representation for Videos,'' IEEE/CVF Conference on Computer Vision and Pattern Recognition (CVPR), 10270-10279, Jun. 2023.
\bibitem{dnerv} Q. Zhao, M. S. Asif, and Z. Ma, ``DNeRV: Modeling Inherent Dynamics via Difference Neural Representation for Videos,'' IEEE/CVF Conference on Computer Vision and Pattern Recognition (CVPR), 2031-2040, Jun. 2023.
\bibitem{dsnerv} Y. Hao, K. Zhihui, Z. Xiaobo, Q. Tie, S. Xidong, and J. Dadong, ``DS-NeRV: Implicit Neural Video Representation with Decomposed Static and Dynamic Codes,'' IEEE/CVF Conference on Computer Vision and Pattern Recognition (CVPR), 23019-23029, Jun. 2024.
\bibitem{pnerv} Q. Zhao, M. Asif, and Z. Ma, ``PNeRV: Enhancing Spatial Consistency via Pyramidal Neural Representation for Videos,'' arXiv preprint arXiv:2404.08921, 2024.
\bibitem{nvrc} H. Kwan, G. Gao, F. Zhang, A. Gower, and D. Bull, ``NVRC: Neural Video Representation Compression,'' arXiv preprint arXiv:2409.07414, 2024.
\bibitem{sb} N. Rahaman, A. Baratin, D. Arpit, F. Draxler, M. Lin, F. Hamprecht, Y. Bengio, and A. Courville, ``On the Spectral Bias of Neural Networks,'' Proceedings of Machine Learning Research, 5301-5310, Jun. 2019.
\bibitem{sb2} Y. Cao, Z. Fang, Y. Wu, D. Zhou, and Q. Gu, ``Towards Understanding the Spectral Bias of Deep Learning,'' arXiv preprint arXiv:1912.01198, 2020.
\bibitem{sb3} Z. Cai, H. Zhu, Q. Shen, X. Wang, and X. Cao, ``Batch Normalization Alleviates the Spectral Bias in Coordinate Networks,'' IEEE/CVF Conference on Computer Vision and Pattern Recognition (CVPR), 25160-25171, Jun. 2024.
\bibitem{finer} Z. Liu, H. Zhu, Q. Zhang, J. Fu, W. Deng, Z. Ma, Y. Guo, X. Cao, ``FINER: Flexible Spectral-bias Tuning in Implicit NEural Representation by Variable-periodic Activation Functions,'' IEEE/CVF Conference on Computer Vision and Pattern Recognition (CVPR), 2713-2722, Jun. 2024.
\bibitem{convnext} Z. Liu, H. Mao, C. Wu, C. Feichtenhofer, T. Darrell, and S. Xie, ``A ConvNet for the 2020s,'' IEEE/CVF Conference on Computer Vision and Pattern Recognition (CVPR), 11976-11986, Jun. 2022.
\bibitem{resnet} K. He, X. Zhang, S. Ren, and J. Sun, ''Deep Residual Learning for Image Recognition,'' IEEE Conference on Computer Vision and Pattern Recognition (CVPR), 770-778, Jun. 2016.
\bibitem{rchnerv} T. Hayami, and H. Watanabe, ``Implicit Neural Representation for Videos Based on Residual Connection,'' arXiv preprint arXiv:2407.06164, 2024. 
\bibitem{davis} F. Parazzi, J. Pont-Tuset, B. McWilliams, L. Van Gool, M. Gross, and A. Sorkine-Hornung,``A Benchmark Dataset and Evaluation Methodology for Video Object Segmentation,'' IEEE Conference on Computer Vision and Pattern Recognition (CVPR), 724-732, Jun. 2016.
\end{thebibliography}
\end{document}